\documentclass[conference]{IEEEtran}
\usepackage{cite}
\usepackage{amsmath}
\usepackage{amssymb}
\usepackage{mathtools}
\usepackage{arydshln}
\usepackage{graphics}

\usepackage{graphicx}
\usepackage{subcaption}

\usepackage{url}
\usepackage{multirow}
\usepackage{ifthen}
\usepackage{tikz}
\usetikzlibrary{shapes, calc, shapes, arrows, positioning}

\makeatletter
\newcommand{\removelatexerror}{\let\@latex@error\@gobble}
\makeatother
\usepackage{algorithm}
\usepackage[noend]{algpseudocode}

\algrenewcommand\algorithmicrequire{\textbf{Input:}}
\algrenewcommand\algorithmicensure{\textbf{Output:}}

\ifCLASSINFOpdf
\else
\fi

\begin{document}
%


\title{Learning from multivariate discrete sequential data using a restricted Boltzmann machine model}

\author{\IEEEauthorblockN{Jefferson Hernandez}
\IEEEauthorblockA{Escuela Superior Politecnica del Litoral (ESPOL)\\
Campus Gustavo Galindo Velasco\\
Guayaquil 09-01-5863, Ecuador\\
Email: jefehern@espol.edu.ec}
\and
\IEEEauthorblockN{Andres G. Abad}
\IEEEauthorblockA{Escuela Superior Politecnica del Litoral (ESPOL)\\
Campus Gustavo Galindo Velasco\\
Guayaquil 09-01-5863, Ecuador\\
Email: agabad@espol.edu.ec}}


%


\IEEEoverridecommandlockouts
\IEEEpubid{\makebox[\columnwidth]{978-1-5386-6740-8/18/\$31.00 \copyright 2018 IEEE \hfill} \hspace{\columnsep}\makebox[\columnwidth]{ }}

\maketitle

\begin{abstract}
A restricted Boltzmann machine (RBM) is a generative neural-network model with many novel applications such as collaborative filtering and acoustic modeling. An RBM lacks the capacity to retain memory, making it inappropriate for dynamic data modeling as in time-series analysis. In this paper we address this issue by proposing the $p$-RBM model, a generalization of the regular RBM model, capable of retaining memory of $p$ past states. We further show how to train the $p$-RBM model using contrastive divergence and test our model on the problem of predicting the stock market direction considering 100 stocks of the NASDAQ-100 index. Obtained results show that the $p$-RBM offer promising prediction potential.
\end{abstract}



%
\IEEEpeerreviewmaketitle

\section{Introduction}
\begin{figure}[h]
	\centering
	\def\layersep{1.5cm} 
	\def\numvis{4} 
	\def\numhid{3} 
	\begin{tikzpicture}[
	node distance=\layersep,
	line/.style={-} 
	]
	\tikzstyle{neuron}=[draw,circle,fill=black!25,minimum size=21pt,inner sep=0pt];
	\tikzstyle{visible neuron}=[neuron, fill=white];
	\tikzstyle{hidden neuron}=[neuron, fill=white];
	\tikzstyle{annot}=[text width=4em];

	\foreach \name / \y in {1,...,\numvis}
	\node[visible neuron] (V\name) at (\y,0) {$v_\y$};

	\foreach \name / \y in {1,...,\numhid}
	\pgfmathparse{\y + (\numvis - \numhid) * 0.5}
	\node[hidden neuron] (H\name) at (\pgfmathresult, \layersep) {$h_\y$};

	\foreach \source in {1,...,\numvis}
	\foreach \dest in {1,...,\numhid}
	\draw[line] (V\source) -- (H\dest);

	\ifthenelse{\numvis > \numhid}
	{
		\node[annot,left of=V1, node distance=1cm] (hl) {Visible};
		\node[annot,above of=hl] {Hidden};
	}
	{
		\node[annot,left of=H1, node distance=1cm] (hl) {Hidden};
		\node[annot,below of=hl] {Visible};
	}
	\end{tikzpicture}
	\caption[RBM Topology]{Example of an RBM  with \numvis{} visible units and \numhid{} hidden units.}
	\label{fig:rbmtopology}
\end{figure}
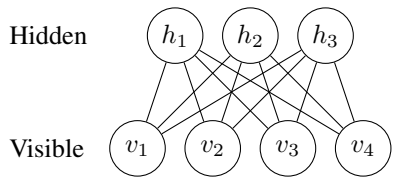

A restricted Boltzmann machine (RBM) is a generative neural-network model used to represent the distribution of random observations. In RBM, the independence structure between its variables is modeled through the use of latent variables (see Figure \ref{fig:rbmtopology}). RBMs were introduced in \cite{smolensky_information_1986}; although it was not until Hinton proposed contrastive divergence (CD) as a training technique in \cite{hinton_training_2002}, that their true potential was unveiled.

Restricted Boltzmann machines have proven powerful enough to be effective in diverse settings. Applications of RBM include: collaborative filtering \cite{salakhutdinov_restricted_2007}, acoustic modeling \cite{dahl_phone_2010}, human-motion modeling \cite{taylor_modeling_2006,taylor_factored_2009}, and music generation \cite{boulanger-lewandowski_modeling_2012}.

Restricted Boltzmann machine variations have gained popularity over the past few years. Two variations relevant to this work are: the RNN-RBM \cite{dahl_phone_2010} and the conditional RBM \cite{taylor_factored_2009}. The RNN-RBM estimates the density of multivariate time series data by pre-training an RBM and then training a recurrent neural network (RNN) to make predictions; this allows the parameters of the RBM to be kept constant serving as a prior for the data distribution while the biases are allowed to be modified by the RNN to convey temporal information. Conditional RBMs, on the other hand, estimate the density of multivariate time series data by connecting past and present units to hidden variables. However, this makes the conditional RBM unsuitable for traditional CD training \cite{mnih_conditional_2012}.

In this paper, we focus on the use of a modified RBM model that does not keep its parameters constant in each time step (unlike the RNN-RBM); and that adds hidden units for past interactions and lacks connections between past and future visible units (unlike the conditional RBM). Our model is advantageous because of two factors: (1) the topology of the model can be changed easily because it is controlled by a single set of parameters; and (2) the structure of our model allow the modeling of auto-correlation within a time series and correlation between multiple time series. These two factors allow many models to be readily tested and compared.

We show the performance of our model by applying it to the problem of forecasting stock market directions, i.e. predicting if the value of a stock will rise or fall after a pre-defined period of time. Previous work on the problem include \cite{huang_forecasting_2005} where a support vector machine model was trained on the  NIKKEI 225 index, the Japanese analog of the Dow Jones Industrial Average index.


The rest of the paper is organized as follows. Section II presents a review of RBMs describing their energy function and training method trough CD. Section III introduces our proposed model, called the $p$-RBM, which can be viewed as an ensemble of RBMs with the property of recalling past interactions. In section IV we apply our model to 100 stocks of the NASDAQ-100 index and show its prediction results. Conclusions and future research directions are provided in Section V.


\section{The Restricted Boltzmann Machine Model}

Restricted Boltzmann machines are formed by $n$ visible units, which we represent  as $\textbf{v} \in \{ 0,1\}^{n}$; and  $m$ hidden units, which we represent as $\textbf{h} \in \{ 0,1\}^{m}$. The joint probability of these units is modeled as
\begin{equation}
\begin{aligned}
p(\textbf{v}, \textbf{h} )=\frac{1}{Z} e^{-  E(\textbf{v}, \textbf{h})};
\end{aligned}
\end{equation}
where the energy function $E(\textbf{v}, \textbf{h})$ is given by
\begin{equation}
\begin{aligned}
E(\textbf{v}, \textbf{h})= -\textbf{a}^{\intercal}\textbf{v} -\textbf{b}^{\intercal} \textbf{h} -\textbf{v}^{\intercal} \textbf{W} \textbf{h};
\end{aligned}
\end{equation}
matrix $\textbf{W} \in \mathbb{R}^{n \times m}$ represents the interaction between visible and hidden units; $\textbf{a} \in \mathbb{R}^{n}$ and $\textbf{b} \in \mathbb{R}^{m}$ are the biases for the visible an hidden units, respectively; and $Z$ is the partition function defined by
\begin{equation}
\begin{aligned}
Z(\textbf{a}, \textbf{b}, \textbf{W})=\sum_{\textbf{v}, \textbf{h}}e^{-  E(\textbf{v}, \textbf{h})} .
\end{aligned}
\end{equation}

The bipartite structure of an RBM is convenient because it implies that the visible units are conditionally independent given the hidden units and vice versa. This ensures that the model is able to capture the statistical dependencies between the visible units, while remaining tractable.

Training an RBM is done via CD, which yields a learning rule equivalent to subtracting two expected values: one with respect to the data and the other with respect to the model. For instance, the update rule for $\textbf{W}$ is
\begin{equation}
\begin{aligned}
\label{eq:update_rule}
 \Delta \textbf{W}=\langle \textbf{v}\textbf{h}^\intercal  \rangle_{\mbox{Data}}-\langle \textbf{v}\textbf{h}^\intercal \rangle_{\mbox{Model}}.
\end{aligned}
\end{equation}
The first term in equation \eqref{eq:update_rule} is the expected value with respect to the data and the second is the expected value with respect to the model. For a detailed introduction to RBMs the reader is referred  to \cite{fischer_training_2014}.

\section{The $p$-RBM Model}

\begin{figure*}[ht!]
\centering
\includegraphics[width=0.65\linewidth]{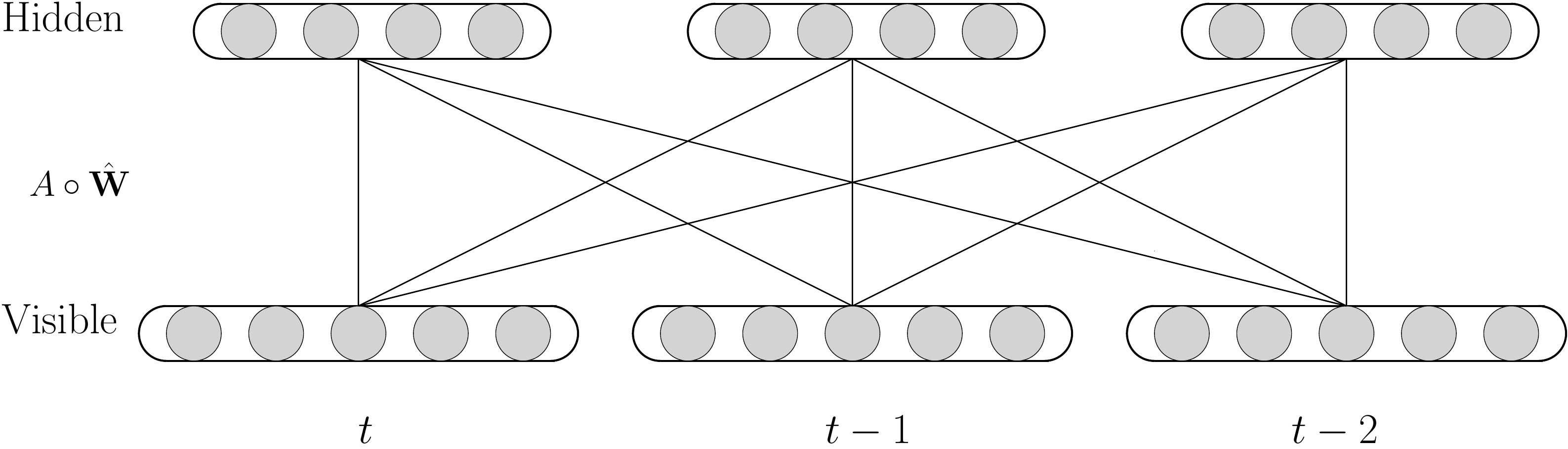}
\caption{Example of a $p$-RBM with $p=3$ steps in the past.}
\label{fig: p_rbm}
\end{figure*}

We generalized the RBM model by constructing an ensemble of RBMs, each one representing the state of the system at $p$ connected moments in time. One way to visualize this is to think of it as $p$ distinct RBMs connected together so as to model the correlation between moments in time. Each RBM contains a representation of the object of interest at different time steps, for example pixels of a video frame, or the value of a dynamic economic index. We then added connections between all the visible and hidden units across time to model their autocorrelation (see Figure \ref{fig: p_rbm}).

Our model resembles a Markov chain of order $p$, because the RBM at time $t$ is conditionally independent of the past, given the previous $p$ RBMs. We showed that even with these newly added time connections the model remains tractable and can be trained in a similar fashion to that of a single RBM.

For convenience, we bundled the visible and hidden units in block vectors denoted $\tilde{\textbf{v}}$ and $\tilde{\textbf{h}}$, respectively; we also included a vector of ones of appropriate size, to account for biases interactions; giving
\begin{equation}
\begin{aligned}
\tilde{\textbf{v}}=\left[\begin{array}{c;{2pt/2pt}c;{2pt/2pt}c;{2pt/2pt}c;{2pt/2pt}c}
\textbf{v}_t &\textbf{v}_{t-1} & \cdots & \textbf{v}_{t-p} & \mathbf{1}
\end{array}\right]^\intercal, \mbox{ and}
\end{aligned}
\end{equation}
\begin{equation}
\begin{aligned}
\tilde{\textbf{h}}=\left[\begin{array}{c;{2pt/2pt}c;{2pt/2pt}c;{2pt/2pt}c;{2pt/2pt}c}
\textbf{h}_t &\textbf{h}_{t-1} & \cdots & \textbf{h}_{t-p} & \mathbf{1}
\end{array}\right]^\intercal.
\end{aligned}
\end{equation}
The parameters can be bundled in a block matrix as follows
\begin{equation}
\begin{aligned}
\label{eq:bmatrix}
\resizebox{0.5\textwidth}{!}{$
\tilde{\textbf{W}}=\left[
\begin{array}{c;{2pt/2pt}c;{2pt/2pt}c;{2pt/2pt}c;{2pt/2pt}c}
\textbf{W}^{v_t,h_t} & \textbf{W}^{v_t,h_{t-1}}&\cdots& \textbf{W}^{v_t,h_{t-p}}& \textbf{W}^{v_t} \\ \hdashline
\textbf{W}^{v_{t-1},h_t} & \textbf{W}^{v_{t-1},h_{t-1}} &\cdots & \textbf{W}^{v_{t-1},h_{t-p}}& \textbf{W}^{v_{t-1}} \\ \hdashline
\vdotswithin{} & \vdotswithin{} &\vdotswithin{} & \vdotswithin{} & \vdotswithin{} \\ \hdashline
\textbf{W}^{v_{t-p},h_t} & \textbf{W}^{v_{t-p},h_{t-1}}&\cdots& \textbf{W}^{v_{t-p},h_{t-p}}& \textbf{W}^{v_{t-p}} \\ \hdashline
\textbf{W}^{h_t} & \textbf{W}^{h_{t-1}} &\cdots & \textbf{W}^{v_{t-p}}& \textbf{0}
\end{array}\right]$}.
\end{aligned}
\end{equation}

We added a new hyperparameter $\textbf{A}\in[0,1]^{p+2,p+2}$ to the model, that acts as a forgetting rate, allowing the model to prioritize connections that are closed in time. While this new hyperparameter could be learned through, for instance, cross-validation or bayesian-optimization methods, in this work it is imposed.

We proposed the following structure for matrix $\textbf{A}$
\begin{equation}
\begin{aligned}
 \textbf{A}= \left\{
 \begin{array}{ll}
 \alpha^{|i-j|} & i,j \leq p  \\
 1 & \mbox{elsewhere} \\
 \end{array}
 \right.
 \end{aligned},
\end{equation}
for fixed $\alpha \in [0,1]$; or expressed in matrix form as
\begin{equation}
\begin{aligned}
\label{eq:Alpha}
\textbf{A}=\left[
\begin{array}{ccccc}
\alpha^{0} & \alpha^{1}&\cdots&  \alpha^{p}& 1 \\
 \alpha^{1} &  \alpha^{0} &\cdots &  \alpha^{p-1}& 1 \\
\vdotswithin{} & \vdotswithin{} &\vdotswithin{} & \vdotswithin{} & \vdotswithin{} \\
 \alpha^{p} &  \alpha^{p-1}&\cdots&  \alpha^{0}& 1 \\
1 & 1 &\cdots & 1& 1
\end{array}\right].
\end{aligned}
\end{equation}
For $\alpha=1$ the model becomes fully connected, with all connections having the same significance. For $\alpha=0$  the model is completely disconnected. Thus, matrix $\textbf{A}$ has some control on the topology of the model. \\

The energy function for our $p$-RBM model is given by
\begin{equation}
\begin{split}
\begin{aligned}
\label{eq:model}
E(\tilde{\textbf{v}}, \tilde{\textbf{h}}) ={}&-\tilde{\textbf{v}}^\intercal (\textbf{A} \circ \tilde{\textbf{W}}) \tilde{\textbf{h}}\\
={} &  -\sum_{i=0}^{p} \sum_{j=0}^{p} \alpha^{|i-j|} \textbf{v}_{t-i}^\intercal \textbf{W}^{v_{t-i},h_{t-j}}\textbf{h}_{t-j}\\
&  -\sum_{i=0}^{p}\textbf{v}_{t-i}^\intercal  \textbf{W}^{v_{t-i}} - \sum_{j=0}^{p} \textbf{h}_{t-j}^\intercal \textbf{W}^{h_{t-j}} ,
\end{aligned}
\end{split}
\end{equation}
where $\circ$ denotes the Hadamard product, indicating element-wise multiplication between matrices.

The energy function given in equation \eqref{eq:model} considers the effect of the previous interaction. It provides the model with the capacity to include the past, allowing for greater flexibility than that of a typical RBM. As a consequence, the model can, for instance, model high-dimensional time series and non-linear dynamical systems, or reconstruct video from incomplete or corrupted versions.

The joint probability distribution under the model is given by the following Boltzmann-like distribution
\begin{equation}
\begin{aligned}
p(\tilde{\textbf{v}}, \tilde{\textbf{h}})=\frac{1}{Z(\tilde{\textbf{W}})} e^{-  E(\tilde{\textbf{v}},\tilde{\textbf{h}})},
\end{aligned}
\end{equation}
where $E(\tilde{\textbf{v}},\tilde{\textbf{h}})$ is given in equation \eqref{eq:model}, and  $Z(\tilde{\textbf{W}})$  is the partition function defined as
\begin{equation}
\begin{aligned}
Z(\tilde{\textbf{W}})=\sum_{\tilde{\textbf{v}}, \tilde{\textbf{h}}} e^{-  E(\tilde{\textbf{v}}, \tilde{\textbf{h}})}.
\end{aligned}
\end{equation}

The model structure induces the following relation of conditional independence
\begin{equation}
\begin{aligned}
 p(\tilde{\textbf{h}}|\tilde{\textbf{v}})= & p(\textbf{h}_t|\tilde{\textbf{v}})  p(\textbf{h}_{t-1}|\tilde{\textbf{v}}) \cdots  p(\textbf{h}_{t-p}|\tilde{\textbf{v}}) .
\end{aligned}
\end{equation}

\subsubsection{Derivative of the log-likelihood of the model}
Given  a single training example $\tilde{\textbf{v}}=\{\textbf{v}_t, \textbf{v}_{t-1}, \cdots, \textbf{v}_{t-p} \}$, the log-likelihood of our model is given by
\begin{equation}
\begin{aligned}
\mbox{ln} \mathcal{L}(\tilde{\textbf{W}} | \tilde{\textbf{v}}) & =  \mbox{ln}  P(\tilde{\textbf{v}} | \tilde{\textbf{W}}) \\& = \mbox{ln} \frac{1}{Z(\tilde{\textbf{W}})} \sum_{\tilde{\textbf{h}}} e^{- E(\tilde{\textbf{v}}, \tilde{\textbf{h}})} \\
	& =  \mbox{ln} \sum_{\tilde{\textbf{h}}} e^{- E(\tilde{\textbf{v}}, \tilde{\textbf{h}})}- \mbox{ln} \sum_{\tilde{\textbf{v}},\tilde{\textbf{h}}} e^{- E(\tilde{\textbf{v}}, \tilde{\textbf{h}})}.
\end{aligned}
\end{equation}
In what follows, we replace $E(\tilde{\textbf{v}}, \tilde{\textbf{h}})$ with $E$.

Let $w$ be a parameter in $\tilde{\textbf{W}}$, then the derivative of the log-likelihood w.r.t. $w$ becomes
\begin{equation}
\begin{aligned}
\label{eq:loglik}
\frac{\partial \mbox{ln} \mathcal{L}(\tilde{\textbf{W}} | \tilde{\textbf{v}})}{\partial w} &= -  \sum_{\tilde{\textbf{h}}}  p(\tilde{\textbf{h}}|\tilde{\textbf{v}}) \frac{\partial E}{\partial w} + \sum_{ \tilde{\textbf{h}}, \tilde{\textbf{v}}}  p(\tilde{\textbf{h}}, \tilde{\textbf{v}}) \frac{\partial E}{\partial w}.
\end{aligned}
\end{equation}
Equation \eqref{eq:loglik} shows that, as with the RBM, the derivative of the log-likelihood of the $p$-RBM can be written as the sum of two expectations. The first term is the expectation of the derivative of the energy under the conditional distribution of the hidden variables given an example $\{\textbf{v}_t, \textbf{v}_{t-1}, \cdots, \textbf{v}_{t-p} \}$ from the training set $\mathcal{T}$. The second term is the expectation of the derivative of the energy under the $p$-RBM distribution \cite{fischer_training_2014}.
Note that equation \eqref{eq:loglik} can be written as
\begin{equation}
\label{eq:update}
\begin{aligned}
\Delta w = \frac{\partial\mbox{ln}\mathcal{L}(w |\mathcal{T} )}{\partial w} = \bigg\langle \frac{\partial E}{\partial w}\bigg\rangle_{\mbox{Data}}-\bigg\langle \frac{\partial E}{\partial w} \bigg\rangle_{\mbox{Model}}.
\end{aligned}
\end{equation}
Equation \eqref{eq:update} is the update step corresponding to the $p$-RBM model and is analogous to equation \eqref{eq:update_rule}.
\subsubsection{Contrastive divergence for the model}
When considering our model, the $k$-step contrastive divergence ($\mbox{CD}_k$) becomes
\begin{equation}
\begin{aligned}
\label{eq:cdk}
\mbox{CD}_k(w, \tilde{\textbf{v}}^{(0)})= - \sum_{\tilde{\textbf{h}}^{(0)}} p(\tilde{\textbf{h}}^{(0)} |\tilde{\textbf{v}}^{(0)} )\frac{\partial E}{\partial w} + \sum_{\tilde{\textbf{h}}^{(k)}} p(\tilde{\textbf{h}}^{(k)} |\tilde{\textbf{v}}^{(k)}) \frac{\partial E}{\partial w},
\end{aligned}
\end{equation}
where a Gibbs chain has been run $k$ times on the second term in order to approximate the expectation of the derivative of the energy under the model, given in equation \eqref{eq:loglik}.
\subsubsection{Training the model}
In order to define a training rule for the model, we propose a way to sample for the CD. The sampling of a block vector of hidden variables can be done from the following block vector of probabilities
\begin{equation}
\begin{aligned}
\label{eq:prob}
\mathbb{P}(\tilde{\textbf{h}}|\tilde{\textbf{v}})=\left[\begin{array}{c;{2pt/2pt}c;{2pt/2pt}c;{2pt/2pt}c}
 p(\textbf{h}_t |\tilde{\textbf{v}}) & \cdots &p(\textbf{h}_{t-p} | \tilde{\textbf{v}}) & \textbf{1} 
\end{array}\right]^\intercal.
\end{aligned}
\end{equation}
To sample a vector of visible variables, we construct a block matrix similar to equation \eqref{eq:bmatrix} as
\begin{equation}
\begin{aligned}
\label{eq:bmatrix2}
\mathbb{P}(\tilde{\textbf{v}}|\tilde{\textbf{h}})=\left[
\begin{array}{c;{2pt/2pt}c;{2pt/2pt}c;{2pt/2pt}c}
 p(\textbf{v}_t |\textbf{h}_t)  &\cdots&  p(\textbf{v}_t |\textbf{h}_{t-p})&  p(\textbf{v}_t |\tilde{\textbf{h}}) \\ \hdashline
 p(\textbf{v}_{t-1} |\textbf{h}_t) &\cdots&  p(\textbf{v}_{t-1} |\textbf{h}_{t-p})&  p(\textbf{v}_{t-1} |\tilde{\textbf{h}}) \\ \hdashline
\vdotswithin{}  &\vdotswithin{} & \vdotswithin{} & \vdotswithin{} \\ \hdashline
 p(\textbf{v}_{t-p} |\textbf{h}_t)&\cdots&  p(\textbf{v}_{t-p} |\textbf{h}_{t-p})&  p(\textbf{v}_{t-p} |\tilde{\textbf{h}}) \\ \hdashline
\textbf{1} &\cdots & \textbf{1} & \textbf{0}
\end{array}\right].
\end{aligned}
\end{equation}

\begin{figure*}[ht!]
        \centering
        \begin{subfigure}[h]{.47\textwidth}
                \includegraphics[width=1\textwidth]{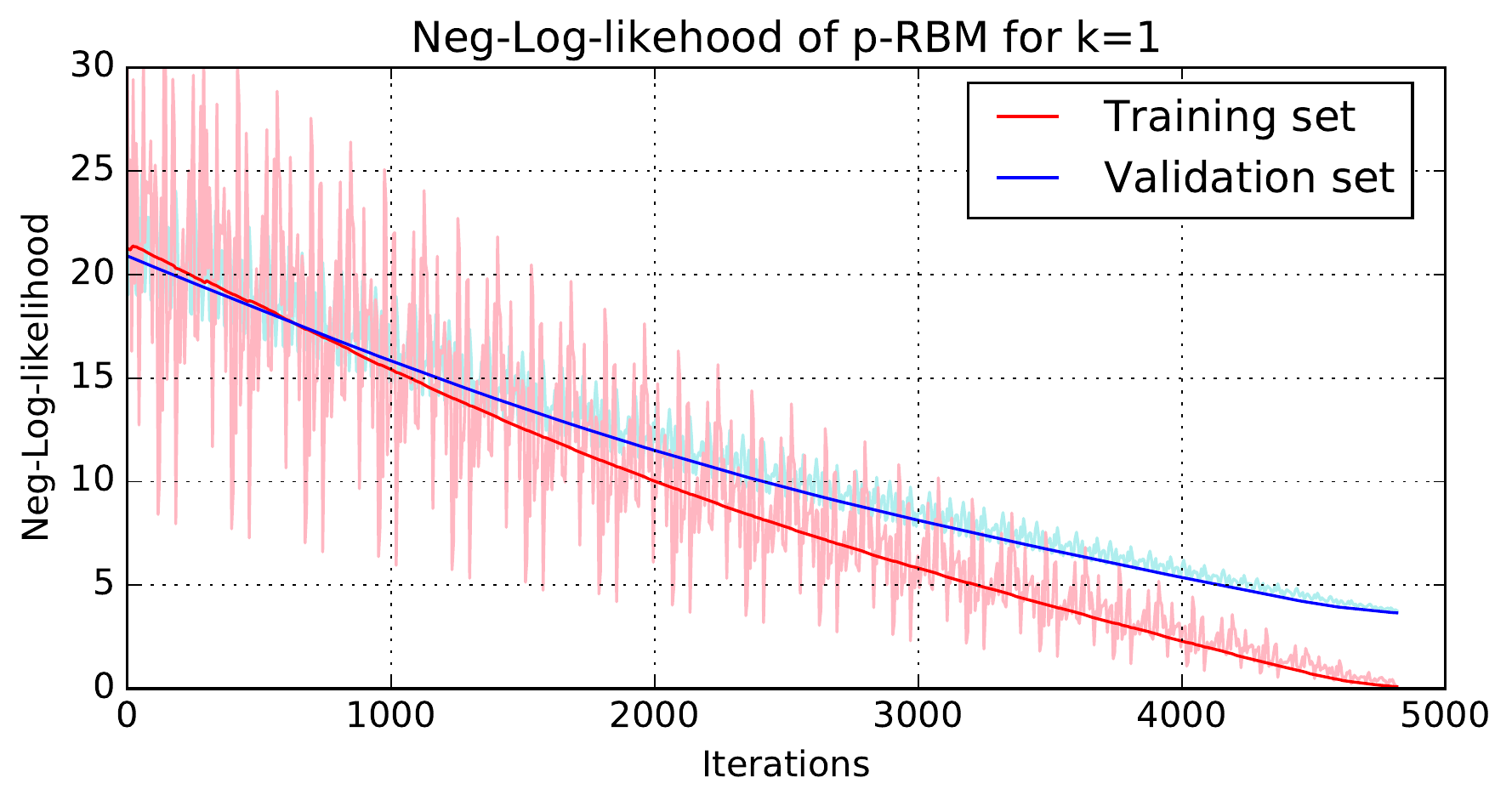}
                \caption{Negative log-likelihood.}
                \label{fig: energy_loss_k_1_a}
        \end{subfigure}
        ~~~ 
        \begin{subfigure}[h]{.47\textwidth}
                \includegraphics[width=1\textwidth]{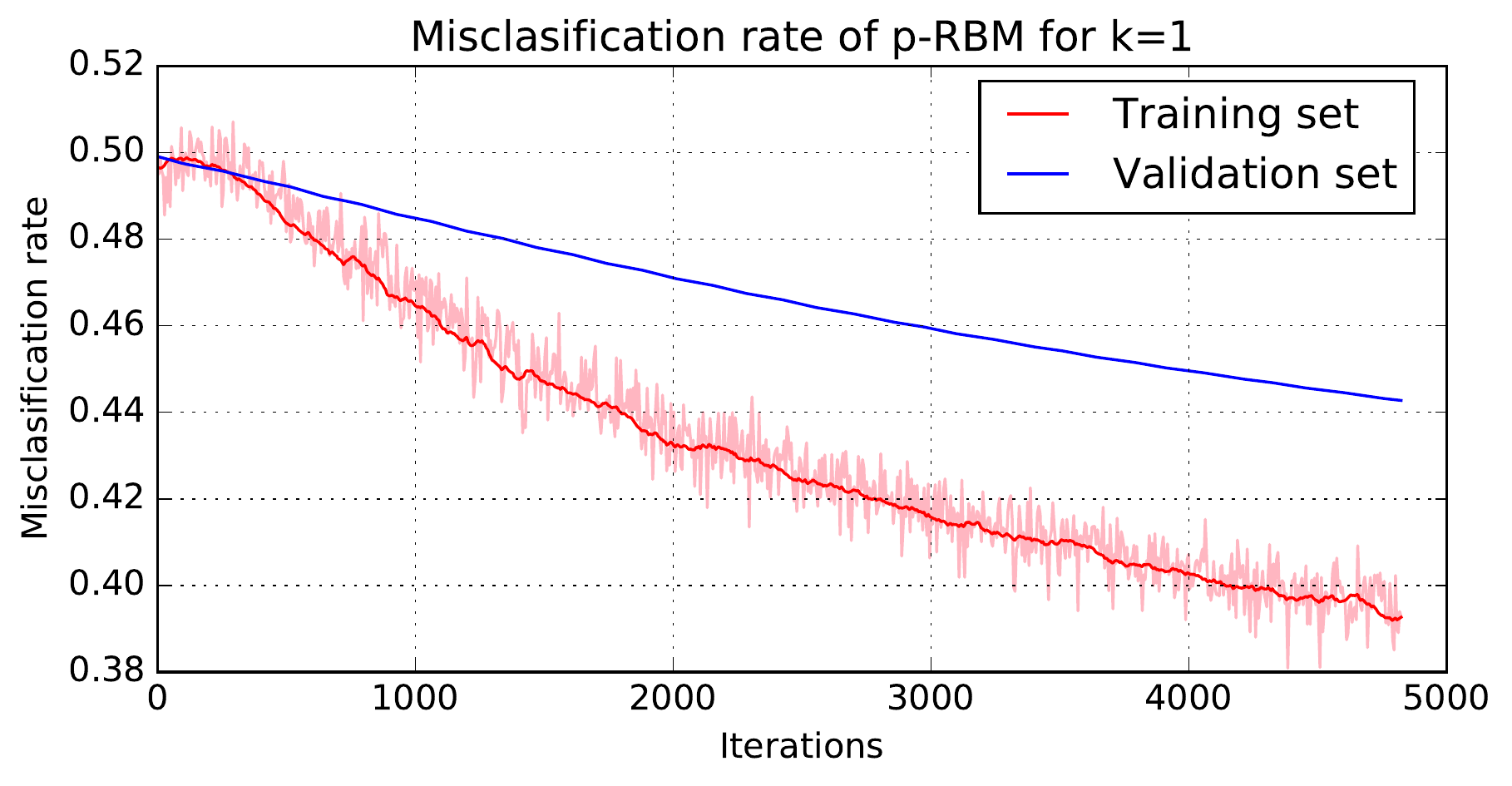}
                \caption{Misclassification rate.}
                \label{fig: energy_loss_k_1_b}
        \end{subfigure}
    \caption{Evaluation of a $p$-RBM model on 100 stocks of the NASDAQ-100 index, with $p=30$, mixing time $k=1$ and $\alpha=0.5$.}\label{fig: energy_loss_k_1}
\end{figure*}
We may obtain the expected value of the hidden vector by repeatedly sampling from equation \eqref{eq:prob}. For our purposes it is better to place them in a block matrix with the same dimension as $\tilde{\textbf{W}}$. We call this matrix $\mathbb{E}[\tilde{\textbf{h}}| \tilde{\textbf{v}}]$ and it is formed by vertically stacking $p+2$ block vectors of the form
\begin{equation}
\begin{aligned}
\label{eq:expec}
\left[\begin{array}{c;{2pt/2pt}c;{2pt/2pt}c;{2pt/2pt}c}
\mathbb{E}[\textbf{h}_t| \tilde{\textbf{v}}] & \cdots &\mathbb{E}[\textbf{h}_{t-p}| \tilde{\textbf{v}}] & \textbf{1} 
\end{array}\right].
\end{aligned}
\end{equation}
The derivatives of $E$ can be written as
\begin{equation}
\begin{aligned}
\label{eq:derivative}
\frac{\partial E}{\partial \tilde{\textbf{W}}}= \textbf{A} \circ \tilde{\textbf{v}} \circ \tilde{\textbf{h}}.
\end{aligned}
\end{equation}
Combining a $\mbox{CD}_k$, as in equation \eqref{eq:cdk}, with equations \eqref{eq:expec} and \eqref{eq:derivative}, equation \eqref{eq:loglik} becomes
\begin{equation}
\begin{aligned}
\label{eq:delta}
\Delta \tilde{\textbf{W}}= \textbf{A} \circ (\tilde{\textbf{v}}^{(0)} \circ \mathbb{E}[\tilde{\textbf{h}}^{(0)}| \tilde{\textbf{v}}^{(0)}]-\tilde{\textbf{v}}^{(k)}\circ \mathbb{E}[\tilde{\textbf{h}}^{(k)}| \tilde{\textbf{v}}^{(k)}]).
\end{aligned}
\end{equation}
The learning rule is then given by
\begin{equation}
\begin{aligned}
\label{eq:learnrule}
\tilde{\textbf{W}}^{(\tau+1)}=\tilde{\textbf{W}}^{(\tau)}+ \eta \Delta \tilde{\textbf{W}},
\end{aligned}
\end{equation}
where $\eta>0$ is the learning rate.

Note that each item in the block matrix can be learned independently, once  the CD step is finalized. This results in a model that is highly suitable for parallelization during training, specially on a GPU where its multicore structure allows the concurrent handling of multiple computational threads.

Algorithm \ref{alg:model} describes the detailed procedure for training the $p$-RBM model.

\begin{algorithm}[H]
\caption{$k$-step contrastive divergence for the $p$-RBM model}
\label{alg:model}
\begin{algorithmic}[1]
\Require{Training set $ \mathcal{T} = \{\textbf{v}_t, \textbf{v}_{t-1}, \cdots, \textbf{v}_{t-p}\}^T_{t=1}$, and a value for $\alpha$}
\Ensure{Updates of all parameters in the model, contained in the block matrix from equation \eqref{eq:bmatrix}}
\ForAll{$\{ \textbf{v}_t, \textbf{v}_{t-1}, \cdots, \textbf{v}_{t-p}\} \in \mathcal{T} $}
	\State $\textbf{v}_t^{(0)},\textbf{v}_{t-1}^{(0)}, \cdots, \textbf{v}_{t-p}^{(0)} \gets  \tilde{\textbf{v}}$
	\For{$k=1,\dots,K-1$}
		\State  sample $\tilde{\textbf{h}}^{(k)} \sim \mathbb{P}(\tilde{\textbf{h}}|\tilde{\textbf{v}}^{(k-1)})$ as in equation \eqref{eq:prob}
		\State  sample $\tilde{\textbf{v}}^{(k)} \sim \mathbb{P}(\tilde{\textbf{v}}| \tilde{\textbf{h}}^{(k)}) $ as in equation  \eqref{eq:bmatrix2}
	\EndFor
	\For{$t=1, \dots, n$}
		\State Calculate  $\mathbb{E}[\tilde{\textbf{h}}| \tilde{\textbf{v}}]$ according to equation  \eqref{eq:expec}
		\State Update $\tilde{\textbf{W}}$ according to equation  \eqref{eq:delta}
	\EndFor
\EndFor
\State \textbf{Return} $\tilde{\textbf{W}}$
\end{algorithmic}
\end{algorithm}

\section{Numerical Results and Analysis}

Stocks in the stock market are correlated (similar sectors influence each other) and autocorrelated (dependent on their previous states) \cite{king_transmission_1990}. This causes the stock market to present itself as chaotic, noisy, unstructured, and non-stationary \cite{abu-mostafa_introduction_1996}. Accurate forecasting of stocks is of interest to organizations trying to make sense of the market data. Traditional approaches to stock-market forecasting include the ARIMA and ARCH models \cite{bontempi_machine_2013,zhang_applying_2009}; while less-traditional approaches include, for instance, hidden Markov models that make use of the partitioning properties to generate fuzzy logic rules  that have been reported to yield better results than those of the ARIMA  model \cite{hassan_combination_2009}.


More recently, machine learning techniques  have been used to predict the stock market. For example, in \cite{gradojevic_non-linear_2007} an artificial neural network (ANN) was combined with fuzzy logic to generate trading strategies, shown to perform better than common buy-and-hold strategies; in \cite{armano_hybrid_2005} various ANNs formed a population of experts that evolves according to genetic programming rules to make forecasts and create investment strategies on the S\&P 500 index.

Likewise, the use of recurrent neural networks (RNNs) is an attractive approach to time-series modeling because they provide a sense of order when handling data, allowing the network to abstract the temporal context of the  series. In \cite{saad_comparative_1998} various neural networks techniques were compared and cases where RNNs outperform feed-forward neural networks were shown.

We trained our model in 100 stocks selected from the NASDAQ-100 index. We chose this set of stocks because most of them belong to the technological sector, a highly interdependent sector. This allows our model to demonstrate its ability to extract hidden relations between stocks.

Our setup consisted on extracting the values of the stocks from a web resource\footnote{\url{https://www.google.com/finance}}, using intervals of 5 minutes during a period of 20 days from June 5, 2017 to June, 25, 2017. The direction of the stock was calculated by subtracting its closing value from its opening value. The obtained set was split in two: 80\% for the training set  and the remaining 20\% for the validation set. We chose the network to remember $p=30$ periods of 5-minute steps. Our model considered a hidden layer with $m=1000$ nodes, $\alpha=0.5$, and $\eta=0.001$. Additionally, we used the mixing rate of $k=1$, since it has been argued in the literature that this choice is enough to appropriately approximate the direction of the gradient \cite{hinton_practical_2012}. Our model was implemented using TensorFlow library version 1.1 \cite{tensorflow2015-whitepaper}.

We studied our model's performance at predicting unseen observations by means of its misclassification error rate, defined as
\begin{equation}
\begin{aligned}
\label{eq:error}
\mbox{Loss}=\frac{1}{2n}\frac{1}{|\mathcal{V}|} \sum_{\tilde{\textbf{v}} \in \mathcal{V}} \sum_{i=1}^{n} |1-\tilde{\textbf{v}}_i f(\tilde{\textbf{v}}_i)|,
\end{aligned}
\end{equation}
where $\mathcal{V}$ is the validation set, $|\mathcal{V}|$ denotes its cardinality, $\tilde{\textbf{v}}_i$ is the real direction of the stock $i$, and $f(\tilde{\textbf{v}}_i)$ is the direction predicted by our model. The loss function used penalizes misclassification cases in the validation set and is solely used for evaluation purposes.

\begin{table*}[ht!]
\centering
\caption{Comparison of errors obtained for different models (less is better).}
\label{table:comparison}
\begin{tabular}{lccccccc}
                      & \multicolumn{5}{c}{\textbf{Iteration}}                                                                                                                             & \multicolumn{1}{l}{}              & \multicolumn{1}{l}{}              \\ \cline{2-8}
\textbf{Model}        & \multicolumn{1}{l}{\textbf{1}} & \multicolumn{1}{l}{\textbf{2}} & \multicolumn{1}{l}{\textbf{3}} & \multicolumn{1}{l}{\textbf{4}} & \multicolumn{1}{l}{\textbf{5}} & \multicolumn{1}{l}{\textbf{Mean}} & \multicolumn{1}{l}{\textbf{Std.}} \\ \hline
\textbf{3 layer LSTM} & 0.4386       & 0.4029                            &  0.4158                   & 0.4495                  &                               0.4432 & 0.4300                                  &  0.0177                                 \\
\textbf{VAR(1)}       & 0.4571                        & 0.4404                      &0.4588                  &0.4582                     &                               0.4587 & 0.4546                                  & 0.0072                                  \\
\textbf{1 layer LSTM} &   0.4542              &   0.4542                 &  0.4572                  &   0.4563                 &                               0.4560 &  0.4556                               & 0.0012                                \\
\textbf{p-RBM}        & 0.4450                    & 0.4945                               &0.4638                             &                               0.4979 &  0.4830                              & 0.4768                                  &  0.0222                                 \\
\textbf{RW}        & 0.4798                         & 0.4790                          & 0.4759                         & 0.5289                         & 0.4780                      & 0.4883                            & 0.0203                            \\
\hline
\end{tabular}
\end{table*}

During training, the negative log-likelihood of the model decreases in the training set and in the validation set (see Figure \ref{fig: energy_loss_k_1_a}); this indicates that the model is learning the distribution of the data. The misclassification rate of our model decreased in the training set and exhibited a cyclic behavior in the validation set. Our model seems to have not overfitted the data; as shown by the fact that the gap between the negative log-likelihood on the training set and on the validation set does not seem to increase over iterations.

We also evaluated the performance of our model by constructing  a confusion matrix on the validation set, shown in TABLE \ref{table:error_table_k_1}.
%
\begin{table}[h]
	\caption{Confusion matrix predicted directions for 100 stocks in the validation set.}
	\label{table:error_table_k_1}
	\centering
	\resizebox{0.30\textwidth}{!}{
		\begin{tabular}{l c c }
			&\multicolumn{2}{c}{\textbf{Predicted direction}}\\
			\cline{2-3}
			\textbf{Real direction} & \textbf{Up} & \textbf{Down}\\
			\cline{1-3}
		    \textbf{Up} & $5,756$ & $6,246$ \\
			\textbf{Down}& $5,057$ & $8,341$ \\
\hline
		\end{tabular}}.
	\end{table} \\
The overall prediction error on the validation set is $0.445$. Considering sequential buy-or-sell decisions, one every 5 minutes, we can construct a simple trading strategy on this set of 100 stocks. Applying the strategy on the validation set---consisting of $25,400$ total trading decisions---resulted in $14,097$ winning and $11,303$ losing trades: a win-to-loss ratio of $1.2472$, and a return of $11 \%$ given that we buyed and sold an equal amount of each stock. More complex trading strategies, based on the obtained predictions, may provide improved results.

\subsection{Comparison to other models}
We compared the forecasting performance of the $p$-RBM model with four other models: a random walk (RW), a vector auto-regression (VAR), and a 1- and 3-layer long short term memory (LSTM).

The RW model produces one-step ahead predictions and is defined as follows
\[
y_{t+1}=y_{t}+\alpha+e_{t},
\]
where $e_t$ is a normal i.i.d. error term and $\alpha$  is the  drift term.

The VAR model is a generalization of the autoregressive model that aims to capture linear interdependencies among several time series. More specifically, a VAR($p$) model is one with $p$ lag if the evolution of the series is a linear function of only their $p$ past values; the forecast is obtained as follows
\[
    \mathbf{y}_t = \mathbf{c} + \mathbf{A}_1 \mathbf{y}_{t-1} + \mathbf{A}_2 \mathbf{y}_{t-2} + \cdots + \mathbf{A}_p \mathbf{y}_{t-p} + \mathbf{e}_t,
\]
where $\mathbf{A}_i$ is a time-invariant  matrix and $\mathbf{e}_t$ is a  vector representing the error term. Here we use a VAR(1) model.

Finally, an LSTM model is a type of recurrent neural network with a series of gates that control information flow within the internal states of the network. This gives the model the ability to remember long and short-term dependencies.

Each of the forecasting models considered here was fitted using the training set. The evaluation procedure was made on the validation set and the relative performance was measured by the misclassification error rate. The results obtained are summarized in TABLE \ref{table:comparison} in descending order from best to worst.

The RW model produced an overall error of $0.4883$, having the worst performance due to, probably, the fact that this model assumes that all the information necessary to predict the future is summarized in the current value only. It also assumes that the increments---up or down---are symmetric, that is, with an expected value of zero. This means that the RW model cannot make predictions different than the current trend.

LSTM comes out first with an overall error of $0.4300$. One reason LSTM performs better than the other methods is its ability to discover hidden structures within the data. Specifically, LSTM may have outperformed the other methods because the data is structurally non-linear. Neither the VAR, that had an overall error of $0.4546$, nor the $p$-RBM, that had an overall error of $0.4768$, can fully capture this non-linearity in the data.

The $p$-RBM comes out in fifth place, just outperforming the RW with drift. We note that the $p$-RBM had a higher variance when compared to the other methods; this might be due to the sampling method used. In fact, we noted that when sampling with a hard-threshold rule, the $p$-RBM achieved a lower variance, but its accuracy worsened.

\section{Conclusion and Future Work}
In this paper we proposed an extension to the RBM model, called the $p$-RBM. We tested our model in the problem of forecasting stock market directions with positive results. Experimentation with real data showed that our model is suitable for problems involving high-dimensional correlated time series. When compared with other models, however, the $p$-RBM showed larger prediction errors than that of 1- and 3-layer LSTM models.

The following are possible research directions for our work: (1) combining our model with reinforcement learning to develop agents that discover optimal trading strategies; (2) discovering the most suitable topology for a problem by exploring the space of hyperparameter matrix $\textbf{A}$; and (3) extending our model to consider continuous visible units.






%

\bibliography{IEEE_RBM}
\bibliographystyle{IEEEtran}


\end{document}